%% file: main.tex
\documentclass[wcp]{jmlr}

\usepackage{longtable}

\usepackage{booktabs}

\usepackage{tabularray}

\usepackage{lineno}

\makeatletter
\let\Ginclude@graphics\@org@Ginclude@graphics 
\makeatother

  \author{\Name{Mohamed Imed Eddine Ghebriout} 
  \Email{im\_ghebriout@esi.dz}\\
  \addr \text{Higher National School of Computer Science - ESI ex INI}
  \AND
  \Name{Halima Bouzidi} \thanks{Corresponding author.}
  \Email{Halima.Bouzidi@uphf.fr}\\
  \addr \text{Université Polytechnique Hauts-de-France, LAMIH/CNRS}
  \AND
  \Name{Smail Niar} \Email{Smail.Niar@uphf.fr}\\
  \Name{Hamza Ouarnoughi} \Email{Hamza.Ouarnoughi@uphf.fr}\\
 \addr \text{Université Polytechnique Hauts-de-France, LAMIH/CNRS, INSA Hauts-de-France}
 }

\definecolor{darkgreen}{RGB}{0,120,0}

\usepackage[super]{nth}

\usepackage{libertine}
\usepackage{wrapfig}

\usepackage{multirow}

\makeatletter
\renewcommand*{\@fnsymbol}[1]{\ensuremath{\ifcase#1\or \dagger \or \dagger\or \ddagger\or
   \mathsection\or \mathparagraph\or \|\or **\or \dagger\dagger
   \or \ddagger\ddagger \else\@ctrerr\fi}}
\makeatother

\usepackage{pifont}

\begin{document}

\title[Harmonic-NAS]{\textbf{Harmonic-NAS}: \underline{H}ardware-\underline{A}wa\underline{r}e \underline{M}ultim\underline{o}dal \underline{N}eural Arch\underline{i}tecture Sear\underline{c}h on Resource-co\underline{n}str\underline{a}ined Device\underline{s}}

\maketitle

\input{Sections/abstract}
\input{Sections/introduction}
\input{Sections/related_work}
\input{Sections/methodology}
\input{Sections/evaluation}
\input{Sections/conclusion}







\bibliography{references}

\input{Sections/appendix}

\end{document}

%% file: Sections/abstract.tex
\begin{abstract}

\noindent The recent surge of interest surrounding Multimodal Neural Networks (MM-NN) is attributed to their ability to effectively process and integrate multiscale information from diverse data sources. MM-NNs extract and fuse features from multiple modalities using adequate unimodal backbones and specific fusion networks. Although this helps strengthen the multimodal information representation, designing such networks is labor-intensive. It requires tuning the architectural parameters of the unimodal backbones, choosing the fusing point, and selecting the operations for fusion. Furthermore, multimodality AI is emerging as a cutting-edge option in Internet of Things (IoT) systems where inference latency and energy consumption are critical metrics in addition to accuracy. In this paper, we propose \textit{Harmonic-NAS}\protect\footnote{https://github.com/Mohamed-Imed-Eddine/Harmonic-NAS}, a framework for the joint optimization of unimodal backbones and multimodal fusion networks with hardware awareness on resource-constrained devices. \textit{Harmonic-NAS} involves a two-tier optimization approach for the unimodal backbone architectures and fusion strategy and operators. By incorporating the hardware dimension into the optimization, evaluation results on various devices and multimodal datasets have demonstrated the superiority of \textit{Harmonic-NAS} over state-of-the-art approaches achieving up to $\sim$\textbf{10.9\%} accuracy improvement, $\sim$\textbf{1.91x} latency reduction, and $\sim$\textbf{2.14x} energy efficiency gain.
\begin{keywords}
Multimodal Learning, Data Fusion, Neural Architecture Search, Edge Computing.
\end{keywords}

\end{abstract}

%% file: Sections/introduction.tex
\section{Introduction} \label{sec:intro}
Our environment is continuously providing us with a broad stream of sensory modalities. Selecting and processing these modalities allows us to take action, react appropriately, and get insights into our surroundings. The term \textit{modality} is used to describe the several forms that sensory information might take (e.g., visual, textual, acoustic) \cite{liang2021multibench}. This \textit{multimodal} perception paradigm has been brought to the sphere of Artificial Intelligence (AI) to bridge further the gap between human brain functioning and neural networks \cite{huang2021makes}. 
Recently, Multimodal Neural Networks (MM-NN) have captivated a lot of interest within the deep learning community since they have proven to be more accurate than their unimodal counterparts in several tasks such as action recognition \cite{molchanov2016online}, image-video captioning \cite{pramanick2021momenta}, sentiment analysis \cite{gong2023circulant}, and healthcare \cite{soenksen2022integrated}.  
However, multimodal neural networks are computation- and memory-demanding. Thus, their deployment on Edge and Tiny devices is constrained by the availability of hardware resources \cite{hou2022analyzing}.

Designing unimodal neural networks is still challenging as it requires tuning a broad set of architectural parameters \cite{elsken2019neural}. The design landscape becomes more complex for multimodal neural networks \cite{perez2019mfas} as it typically involves various backbones and fusion networks for unimodal and multimodal feature selection and extraction, respectively. Each unimodal backbone and fusion network is characterized by a specific set of architectural parameters and assigned a particular role in the multimodal learning scheme. Furthermore, incorporating the hardware dimension into the design process of multimodal networks limits their ability to fuse a large quantity of information \cite{rashid2023tinym2net}. 

Recently, Neural Architecture Search (NAS) \cite{elsken2019neural} has emerged as a data-driven approach to automate the design of neural networks by searching for the optimal set of architectural parameters within a predefined search space. 
Typical  NAS approaches adopt evolutionary algorithms \cite{jian2023eenas} or differentiable architecture search \cite{liu2018darts} as a search strategy. 
With the emergence of Edge-AI, Hardware-aware NAS \cite{cai2018proxylessnas} also added hardware efficiency (e.g., latency, energy, memory) as an optimization objective. The NAS paradigm has been leveraged for multimodal networks since MFAS \cite{perez2019mfas} and MM-NAS \cite{yu2020deep} in which the fusion architecture is searchable for visual-textual modalities. BM-NAS \cite{yin2022bm} provides a more general framework to jointly search for the fusion architecture and operators. However, related works have yet to investigate making the MM-NN fully searchable -through unimodal backbones and multimodal fusion networks- across different modalities, tasks, and datasets. Furthermore, the hardware dimension still needs to be included in existing multimodal-NAS frameworks to ease the deployment on resource-constrained devices. 

\subsection{Novel Contributions}
In this paper, we present \textit{Harmonic-NAS}, a novel Hardware-aware NAS framework for the design of Multimodal Neural Networks on resource-constrained Edge devices. Our proposed framework encompasses the following novelties and contributions:
\begin{enumerate}
    \item \textit{Harmonic-NAS} co-optimizes the design of unimodal backbones and fusion networks to learn an effective joint embedding of features from multiple modalities.
    \item We make the MM-NN fully searchable through a hierarchical search space for \textbf{(i)} Unimodal backbones, built upon the once-for-all supernets \cite{cai2019once} and \textbf{(ii)} Multimodal fusion networks, built upon the differentiable search space of DARTS \cite{liu2018darts}.
    \item To solve the bi-level design space exploration problem, \textit{Harmonic-NAS} includes a two-tier optimization, where the first search stage is an evolutionary algorithm for unimodal backbone networks, whereas the second search stage is a differentiable NAS for fusion networks, with an integrated hardware-related loss function in both search stages.
    \item We demonstrate the efficiency of \textit{Harmonic-NAS} by conducting experiments on various multimodal datasets and Edge devices. Empirical results have seen up to $\sim$\textbf{10.9\%} accuracy improvement, $\sim$\textbf{1.91x} latency reduction, and  $\sim$\textbf{2.14x} energy gain, stipulating further the importance of the hierarchical design optimization for multimodal NNs on Edge devices.
\end{enumerate}

%% file: Sections/related_work.tex
\section{Related Work}\label{sec:sota}

\subsection{Multimodal Neural Networks}
The multimodality paradigm involves feature fusion from multiple modalities to learn a joint embedding of global information. Initially,  fusion approaches operate on the extreme levels of feature abstraction within the neural network on early layers with low-level features and on the last layers with high-level features. Early fusion operates on an input level, whereas late fusion operates on an output level using aggregation operators such as averaging or voting. 
As modern unimodal backbones are deeper and larger with features extracted at many levels, intermediate fusion has been introduced to provide more flexibility in the fusion position by operating on the intermediate feature-map level \cite{vielzeuf2018centralnet}. 
However, one challenge arises in determining the fusion placement as dense fusion networks \cite{yu2020deep} fail to scale when the unimodal backbone networks deepen, resulting in an exponential increase of the fusion parameters. Fused data and fusion operators define the joint embedding granularity and quality. Simple features can be fused using sum, concatenation, or tensor operations \cite{liu2018efficient}. Nevertheless, for complex multimodal tasks, sophisticated fusion networks such as Attention \cite{nagrani2021attention}, Graph Neural Networks \cite{cai2022multimodal}, and Mixture-of-experts \cite{mustafa2022multimodal} are needed to effectively learn the complex interactions between modalities. 

\subsection{Neural Architecture Search}
Neural Architecture Search (NAS) aims to automate the design exploration of neural networks \cite{elsken2019neural}. The NAS is typically viewed as a black-box optimization taking as input the search space and the optimization objective (e.g., accuracy). 
As a search strategy, existing NAS frameworks generally employ evolutionary \cite{bouzidi2023hadas, odema2023magnas, jian2023eenas}, or differentiable \cite{liu2018darts} search algorithms. 
However, the NAS process is labor intensive, requiring many training-validation trials on the explored neural networks. 
To alleviate this problem, progressive shrinking \cite{liu2018progressive} and once-for-all (OFA) supernets \cite{cai2019once} have been proposed. The once-for-all scheme is widely adopted for one-shot NAS \cite{dong2019one}, which consists of training a supernet comprising all the NN candidates once via weight-sharing and reusing the pretrained supernet to sample NNs during the search. While traditional NAS paradigms assume a discrete encoding of the NN architecture, Differentiable NAS (DARTS) \cite{liu2018darts} proposed a continuous relaxation of the NN encoding, allowing for gradient-based optimization. The NAS paradigm has been incorporated first by MFAS \cite{perez2019mfas} to serve multimodal NNs. However, MFAS operates on a priori fixed backbones and only uses concatenation as a fusion operator while searching for the fusion positions. MM-NAS \cite{yu2020deep} then followed up by refashioning the MM-NN into an encoder-decoder scheme with fully searchable fusion operators. Still, the unimodal backbones in MM-NAS are highly specialized and lack scalability to other types of neural networks. MUFASA \cite{xu2021mufasa} has first attempted to jointly optimize the unimodal backbones and the fusion network using an evolutionary NAS and a Transformer backbone on the MIMIC-CCS dataset \cite{johnson2016mimic}. Nevertheless, MUFASA targets one small dataset with a relatively simple search space. 
Moreover, their one-stage evolutionary search is not scalable to sophisticated backbones, fusion operators, and multimodal tasks. 
The recent BM-NAS framework \cite{yin2022bm} provides a more general and scalable fusion search using differentiable NAS \cite{liu2018darts}. Nevertheless, BM-NAS operates on fixed unimodal backbones, overlooking opportunities for further performance gains from optimizing the unimodal feature extraction. Additionally, the hardware dimension is still missing in existing multimodal NAS approaches for further deployment on resource-constrained devices.

\subsection{Hardware Acceleration for Multimodal Neural Network}
Modern IoT systems (e.g., wearable devices) comprise sensors gathering data from multiple modalities (e.g. image, audio, text). Processing these data requires over-parameterized multimodal networks with high computation and memory demands. Commodity Tiny and Edge devices have limited resources, burdening further the deployment of MM-NNs \cite{rashid2023tinym2net}. Thus, hardware-aware optimization is needed when designing MM-NNs. Hardware-aware NAS approaches \cite{cai2018proxylessnas,wu2019fbnet} have contemplated integrating the hardware dimension into the NAS process for unimodal neural networks. Nonetheless, the multimodal case is still understudied. While few attempts have been made towards the optimization of MM-NN on Tiny and Edge devices by integrating latency as objective \cite{liu2021searching}, exploiting computation parallelism \cite{zhang2022h2h} and mixed-precision quantization \cite{rashid2023tinym2net}, a holistic design exploration framework for \texttt{HW}$\times$\texttt{MM-NN} is lacking in existing works.

\begin{table}[ht]
\centering
\caption{Comparison between existing works on Multimodal-NAS and ours.}
\vspace{1ex}
\fontsize{10}{10}\selectfont
\scalebox{.9}{
\label{tab:comparison}
\begin{tabular}{lcccccc}
\hline
Multimodal-NAS work &
  \multicolumn{1}{l}{MFAS} &
  \multicolumn{1}{l}{MM-NAS} &
  \multicolumn{1}{l}{MUFASA} &
  \multicolumn{1}{l}{MMTM} &
  \multicolumn{1}{l}{BM-NAS} &
  \multicolumn{1}{l}{\textbf{Harmonic-NAS (ours)}} \\ \hline
Unimodal backbone search        &   &            & \checkmark &   &   & \textbf{\checkmark} \\
Unimodal feature selection       & \checkmark  &            & \checkmark & \checkmark & \checkmark & \textbf{\checkmark} \\
Fusion micro-arch. search & \checkmark & \textbf{\checkmark} & \checkmark & \checkmark & \checkmark & \textbf{\checkmark} \\
Fusion macro-arch. search &   &            & \checkmark &   &   & \textbf{\checkmark} \\
Multimodal design flexibility    &   &            &   &   & \checkmark & \textbf{\checkmark} \\
Multimodal tasks scalability     &   & \checkmark          &   & \checkmark & \checkmark & \textbf{\checkmark} \\
Hardware awareness               &   &            &   &   &   & \textbf{\checkmark} \\ \hline
\end{tabular}
}
\end{table}

\noindent Previous multimodal-NAS approaches in MFAS \cite{perez2019mfas}, MM-NAS \cite{yu2020deep}, MUFASA \cite{xu2021mufasa}, MMTM \cite{joze2020mmtm}, and BM-NAS \cite{yin2022bm} mostly rely on a priori fixed pretrained backbones as unimodal feature extractors and search only for the fusion network. However, this approach incurs the following drawbacks: 
\textbf{(i)} The a priori fixed unimodal backbones are not initially designed to serve multimodal networks. 
\textbf{(ii)} The Multimodal fusion performance depends highly on the quality of the extracted unimodal features.
\textbf{(iii)} The overlooked performance and efficiency gains that can be obtained from specializing the multimodal fusion to less hardware-demanding backbones.
Thus, \textit{\text{Harmonic-NAS}} aims to make the MM-NN fully searchable by hierarchizing the architectural parameters related to the unimodal backbones and multimodal fusion networks design and optimizing them at once through a two-tier optimization strategy. \textit{\text{Harmonic-NAS}} also includes the hardware dimension as an optimization objective to ease the deployment of MM-NNs on Edge devices that characterize modern IoT systems. In Table \ref{tab:comparison}, we highlight the key differences between related works on multimodal-NAS and ours.

%% file: Sections/methodology.tex
\section{Methodology}\label{sec:method}

\begin{figure}[tp]
\centering
    \includegraphics[width=1.\textwidth]{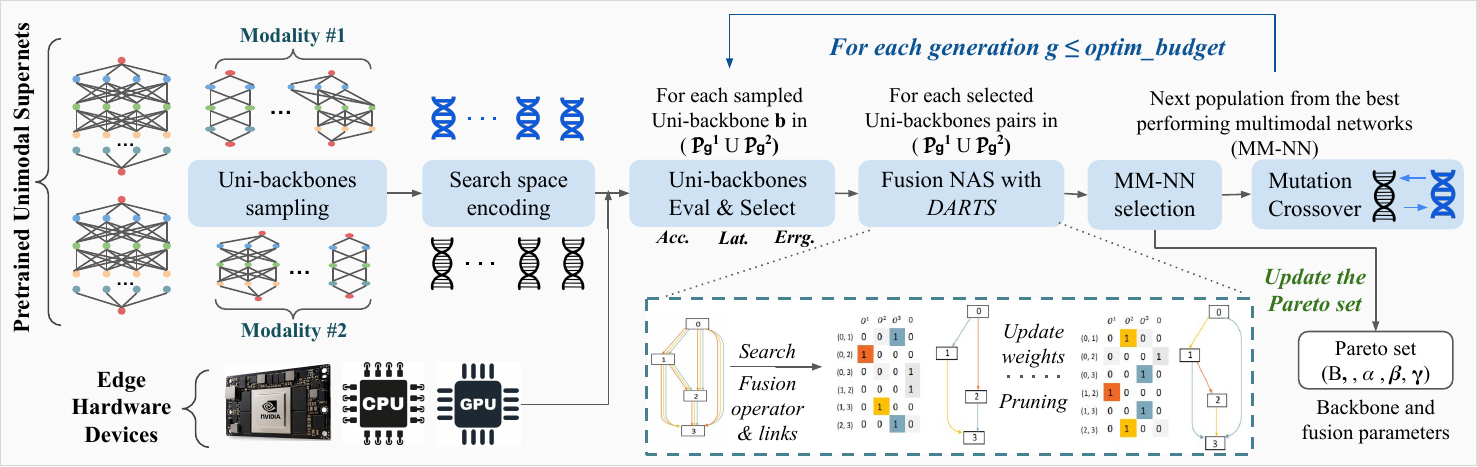}
    \caption{An overview of \textit{Harmonic-NAS} for multimodal neural architecture search.} 
    \label{fig:framework}
\end{figure}

In this paper, we propose \textit{Harmonic-NAS}, a novel framework that aims to make the design of MM-NN fully searchable.
As illustrated in Figure \ref{fig:framework}, \textit{Harmonic-NAS} comprises a two-tier optimization stages:
\textbf{(i)} The \textit{first-stage} searches for optimal unimodal backbones for each \textit{modality}. This involves exploring a search space of modality-specific backbone architectures and evaluating their performance on the target task and their hardware efficiency on the Edge device. \textbf{(ii)} Within this process, the most promising unimodal backbones are selected for fusion network optimization in the \textit{second-stage} to derive multimodal networks. The aim of the \textit{second-stage} is to find a fusion network capable of effectively learning a joint embedding of features, thereby fully leveraging data's intrinsic multimodality. 
\textit{Harmonic-NAS} addresses this challenge by leveraging differentiable NAS (DARTS) to adapt the fusion network to the selected unimodal backbones. 
The two search stages are executed iteratively until reaching a final optimization budget (e.g., evolutionary generations).

\subsection{First-stage: Unimodal Backbone Search}

\subsubsection{Unimodal backbone design and training}
Given the heterogeneity of modalities, tasks, and hardware devices, a meticulous design of unimodal backbones is essential to achieve the optimal performance-efficiency tradeoff. To achieve this, \textit{Harmonic-NAS} adopts a once-for-all approach by designing a \textit{Supernet} comprising diverse neural architecture configurations. 
This will save valuable time for the unimodal backbones search step and facilitate the discovery of more efficient and task-specific NN architectures.

\begin{figure}[tp]
\centering
    \includegraphics[width=\textwidth]{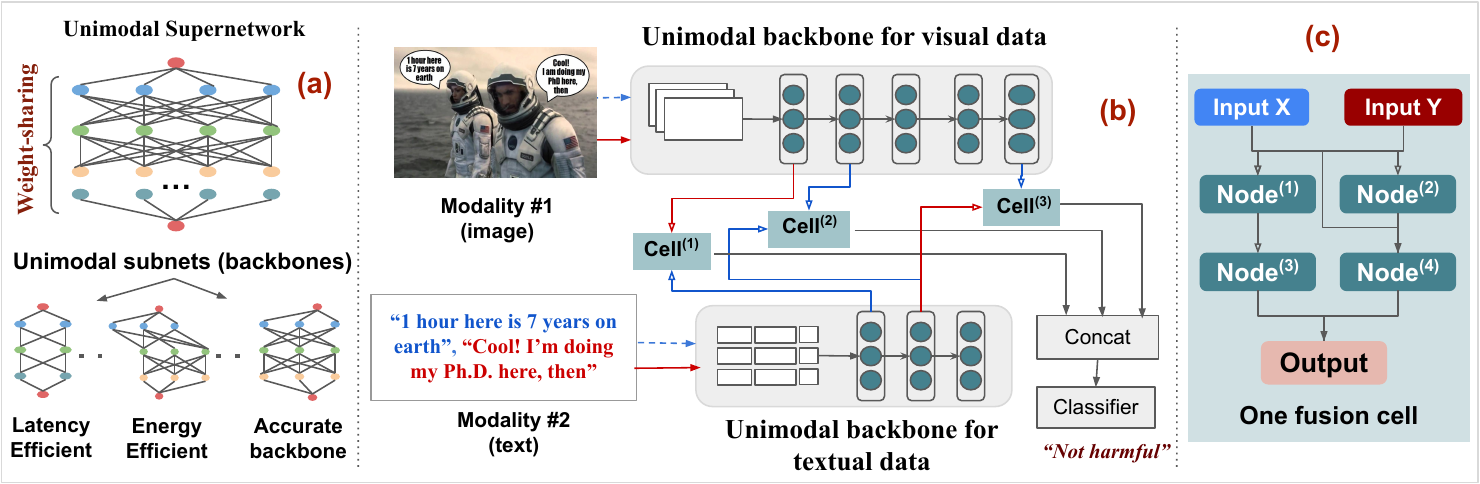}
    \caption{\textbf{(a)} An illustration of the unimodal supernet with subnets sharing weights and neural macro-architecture. \textbf{(b)} A high-level overview of the multimodal NN with modality-specific backbones and fusion network comprises multiple \textit{cells}. \textbf{(c)} A detailed view of one fusion \textit{cell} with multiple nodes, each assigned a particular fusion operator.} 
    \label{fig:networks}
\end{figure}

\noindent{\large \textcircled{\small 1}} \textbf{Supernet Design Specifications:}
A modality-specific supernet $\mathcal{S}_i$ is defined as a hyper-network of subnets sharing the same macro-architecture (i.e., neural blocks) and weights as depicted in Figure \ref{fig:networks}-(a). 
A unimodal backbone $\mathcal{B}_{i}(\cdot)$ (i.e., subnet from $\mathcal{S}_i$) for the $i^{\text{th}}$ modality is represented as a succession of $m$ neural blocks each comprising a sequence of layers $\mathcal{L}$ as follows:
\begin{equation}
    \mathcal{B}_i(\cdot) = \mathcal{B}_i^{m} \circ \mathcal{B}_i^{m-1} \circ ... \circ \mathcal{B}_i^{1} \quad | \quad \mathcal{B}_i^{j} = \mathcal{L}^{d_j} \circ \mathcal{L}^{d_{j}-1} \circ ... \circ \mathcal{L}^{1} \quad | \quad \text{for each} \quad \mathcal{B}_i \in \mathcal{S}_i
    \label{eqn:uni_backbone} 
\end{equation}
Each block $\mathcal{B}_i^{j}$ is characterized by a unique micro-architecture parameterized by a dynamic depth $d_j$, kernel sizes $k_j$, and channel expand ratio $e_j$. As we target deploying MM-NN on resource-constrained devices, we leverage the same search space based on the MobileNet-v3 baseline in \cite{cai2019once}. We note that the backbone search space for the $i^{\text{th}}$ modality is designated as $\mathcal{S}_i$.

\noindent{\large \textcircled{\small 2}} \textbf{Supernet Training:}
Training a \textit{Supernet} can be a challenging task. It differs from training a single NN since we jointly optimize the shared weights of all the subnets as follows:
\begin{equation}\small
\label{eq:general_supernet_obj}
    \min_{\mathbb{W}} \sum_{subnet_i} \mathcal{L}_{test} \big( K(\mathbb{W}, subnet_i), \mathcal{Y} \big)
\end{equation}
where $\mathbb{W}$ denotes the shared weights from the supernet for the architectural configuration $subnet_i$ obtained by a sampling method denoted $K$, and $\mathcal{Y}$ is the ground truths labels. The aim is to optimize $\mathbb{W}$ for each sampled subnet while minimizing the cost of the independent training of subnets via the weight-sharing and adequate sampling scheme that maintains fairness across subnets.

We train our unimodal supernets using \textit{knowledge distillation} from the largest subnet (i.e., teacher) to guide the training of smaller subnets (i.e., students) \cite{wang2021alphanet}. By distilling the knowledge from the largest subnet, we aim to transfer its learned unimodal feature representations to smaller subnets. We also employ the sandwich rule \cite{yu2019universally} to sample subnets at each epoch. This rule involves sampling two types of subnets -- those following a random distribution $\Gamma$ over the search space $\mathcal{S}_i$ and those belonging to a presdefined set $\mathcal{J}$ from $\mathcal{S}_i$. Alternatively, we define the overall set of unimodal subnets evaluated at each epoch as follows: 
\begin{equation} 
\{\mathcal{B}_{im} \vert \mathcal{B}_{im} \sim \Gamma (\mathcal{S}_i) \}_{\ m=1}^{\ \lvert \mathcal{S}_i \rvert} \cup  
\{\mathcal{B}_{in} \vert \mathcal{B}_{in}  \in \mathcal{J}  \}_{\ n=1}^{\ \lvert \mathcal{S}_i \rvert}
\end{equation} 

Therefore, by considering our selection method, the training process of the unimodal supernet aims to optimize the shared weights $\mathbb{W}$ by choosing a uniform distribution of subnets $\Gamma$ in addition to the max-subnet and the min-subnet in $\mathcal{J}$. Our supernet training loss is depicted in equation \ref{eq:supernetçtrain_loss}.

\begin{equation} 
	\small
	\begin{aligned}
		 \underset{\mathbb{W}}{\mathrm{arg\,min}} \, \biggl(
		\sum_{\mathcal{B}_{im} \in \Gamma(\mathcal{S}_i)} \mathcal{L} \left(\mathcal{B}_{im}, \mathbb{W}, \mathcal{Y} \right) + \sum_{\mathcal{B}_{in} \in \mathcal{J}} \mathcal{L} (\mathcal{B}_{in}, \mathbb{W}, \mathcal{Y}) \biggl) \\
\end{aligned} 
\label{eq:supernetçtrain_loss}
\end{equation} 

\subsubsection{Evolutionary search strategy}

Once the unimodal supernets are fully trained, \textit{Harmonic-NAS} employs an evolutionary search strategy at the \textit{first-stage} to explore the design spaces of backbones. The evolutionary search is set to be run for a specific number of generations. For each generation, it creates populations $\mathcal{P}_g^{\mathcal{S}_i}$ of backbones - \textit{for each modality} - from which the multimodal networks will be procured. A value-encoding step is also used to create discrete vectors characterizing the architectural parameters of the neural blocks (i.e., depth, kernel size, and channel expand ratio) of the sampled backbones.

Afterward, each backbone undergoes an evaluation to assess its unimodal performance on the validation set and its hardware metrics of the inference (i.e., latency and energy consumption). The hardware metrics are directly computed using device-specific lookup tables (LUTs). For each backbone $\mathcal{B}_i \in \mathcal{P}_g^{\mathcal{S}_i}$, the evaluation function of the unimodal performance is defined as follows:
\begin{equation}
    \mathcal{U}(\mathcal{B}_i) = Eval\_score(Acc(\mathcal{B}_i), Lat(\mathcal{B}_i), Enrg(\mathcal{B}_i))
\end{equation}

To compute $Eval\_score$ in the underlying multi-objective context, \textit{Harmonic-NAS} incorporates a Pareto ranking using the non-dominated sorting algorithm on the unimodal performance metrics (i.e., accuracy, latency, and energy) and a crowding distance that measures the diversity of backbones in the objective space \cite{deb2000fast}. The evaluation score is then used to identify a subset of the top-performing unimodal backbones $\mathcal{P'}_{g}^{\mathcal{S}_i} \subset \mathcal{P}_{g}^{\mathcal{S}_i}$ on which the second search stage for fusion network will be performed to derive optimal multimodal networks. 

After the completion of the fusion search on $\mathcal{P'}_{g}^{\mathcal{S}_i}$, \textit{Harmonic-NAS} proceeds, through the evolutionary search engine, to the creation of the next population $\mathcal{P}_{g+1}^{\mathcal{S}_i}$ of unimodal backbones using \textit{mutation} and \textit{crossover}. Within this process, a second selection criterion is set only to pick backbones achieving the best performance in their multimodal variant (i.e., with fusion network) as elite solutions for \textit{mutation} and \textit{crossover}. A uniform mutation is employed on the neural block level of backbones by sampling new depth, kernel size, and channel expand ratio under a probability threshold of 0.4. The crossover is applied by randomly picking two unimodal backbones -for the same modality- and swapping their neural blocks under a probability threshold of 0.8.

\subsection{Second-stage: Multimodal Fusion Search} \label{second_stage}
 
\subsubsection{Fusion network search space} \label{fusion_network_search_space}
In light of the scalability of \textit{Harmonic-NAS} to the multimodal task complexity and diversity of backbone architectures, it's crucial to define a generalized search space with all the possible options for the fusion macro-architecture (i.e., number of fusion cells and nodes)
and micro-architecture (i.e., fusion positions and operators). To this end, we adopt a cell-based fusion search space as introduced in \cite{yin2022bm} with a parameterized fusion macro-architecture. As depicted in Figure \ref{fig:networks}-(b), the fusion network is built upon \textit{fusion cells}, wherein each cell selects unimodal features to be fused. These unimodal features can be chosen from the output of intermediate blocks of the backbones or the outputs of previous fusion cells. A fusion cell comprises multiple fusion nodes, as shown in Figure \ref{fig:networks}-(c), each performing a specific fusion operator on the selected features.

\noindent{\large \textcircled{\small 1}} \textbf{Unimodal Feature Selection}:   
Assuming the availability of $m$ modality each processed by a specific backbone $B_1,\dots, B_m$, respectively. A unimodal feature extracted by the $jth$ neural block of the $ith$ backbone is denoted as $B_{\ i}^{(j)}$. The unimodal feature selection procedure consists of choosing for each fusion $\text{Cell}^{(p)}$, two input features from the unimodal features set $\mathcal{F}_1$ as given by (\ref{eq:uni_feature_set}), enabling both inter and intra-modality fusion:
\begin{equation}
\mathcal{F}_1 =[ B_1^{(1)}, ..., B_1^{(N_{B_1})},B_m^{(1)},...,B_m^{(N_{B_m})},\text{Cell}^{(1)},..., \text{Cell}^{(p-1)} ]
\label{eq:uni_feature_set}
\end{equation}
By performing a continuous relaxation on our search space, each fusion $\text{Cell}^{(p)}$ receives a weighted sum of the $\mathcal{F}_1$ elements with their corresponding probabilities $(\alpha)$ of being selected or not. These probabilities are updated during the training phase of the fusion network using DARTS. Then, each fusion $\text{Cell}^{(p)}$ operates on the received weighted sum of input features as follows:
\begin{equation}
Cell^{\ (p)} = \sum\limits_{k=0}^{\ \lvert \mathcal{F}_1 \rvert}\  \overline{a}^{(k,p)}\ (\mathcal{F}_1[k]) \quad , \quad
 \overline{a}^{(k,p)}(s) = \sum\limits_{a \in \mathcal{A}} \frac{\text{exp}(\alpha_a^{(k,p)})} {\sum\limits_{a' \in \mathcal{A}} \text{exp}( \alpha_{a'}^{(k,p)} ) } a(s) 
\label{eq:darts_one}
\end{equation}
Here $\mathcal{A}$ represents a set of two functions: Identity ($o(x)=x$) and Zero ($o(x)=0$). At the evaluation phase, only the input features $(X, Y)$ depicting the highest probabilities will be chosen as unimodal features as follows :
\begin{equation}
(X, Y) = \underset{p<r<k,\ a \in \mathcal{A}}{\mathrm{arg\,max}}(\alpha_a^{(p,k)} \cdot \alpha_a^{(r,k)})
\end{equation}

\noindent{\large \textcircled{\small 2}} \textbf{Multimodal Feature Selection and Fusion Operators}:
At this stage, we investigate the design of the inner structure of one fusion $\text{Cell}^{(p)}$. A fusion cell constitutes $\mathcal{D}$ fusion nodes, each assigned a particular fusion operator from our predefined fusion operators set $\mathcal{FP}$ (See Table \ref{tab:fusion_ops}).
A fusion $\text{Node}^{(d)}$ operates on two inputs -- In the case of the first fusion node, the inputs are directly $\text{Cell}^{(p)}$'s inputs. However, for subsequent fusion nodes, the inputs can also be the outputs of previous fusion nodes. 
More formally, the inputs of a fusion $\text{Node}^{(d)}$, are selected from the multimodal features set $\mathcal{F}_2$ defined as follows:
\begin{equation}
\mathcal{F}_2 =[ X, Y, \text{Node}^{(1)},..., \text{Node}^{(d-1)} ]. \notag
\label{eq:multi_feature_set}
\end{equation}
where $(X, Y)$ are the inputs of the current fusion $\text{Cell}^{(p)}$ while $\text{Node}^{(1,\dots,d-1)}$ denotes the output of the underlying fusion node within $\text{Cell}^{(p)}$.
Similarly to the unimodal feature selection mechanism for the fusion cell, fusion nodes follow the same strategy to select their input features $(x,y)$, as shown in equation \ref{eq:darts_one} expect that here we use the $\beta$ weights instead of $\alpha$ and select inputs from $\mathcal{F}_2$.

Another layer of complexity is added by searching further which fusion operator to be used at each fusion node. This is done by assuming $(x,y)$ as $\text{Node}^{(d)}$ inputs and applying a continuous relaxation over the fusion operators set $\mathcal{FP}$ as follows:
\begin{equation}
\quad
\overline{f}^{(d)}(x, y) = \sum\limits_{f \in \mathcal{FP}} \frac{\text{exp}(\gamma_{f}^{(d)})} {\sum\limits_{f' \in \mathcal{FP}} \text{exp}( \gamma_{f'}^{(d)} ) } f(x, y)
\end{equation}
where $\gamma$ is the weight matrix that sets a priority score for each fusion operator to be selected for each node. At the evaluation phase, the following criterion is used to select the best fusion operator from the fusion set $\mathcal{FP}$:
\begin{equation}
    f^{\ (d)} =  \underset{f \ \in \ \mathcal{FP} }{\mathrm{arg\,max}} \  \gamma_{\ f}^{\ (d)}
\end{equation}

\noindent To further ensure the multimodal fusion effectiveness across various designs of backbones, we consider hardware efficiency as a criterion when defining $\mathcal{FP}$. We select fusion operators that minimize the hardware burden while ensuring effective multimodal fusion. Table \ref{tab:fusion_ops} provides details on the employed fusion operators with a brief explanation of each operator:

\begin{table}[ht!]
\centering
\caption{The set $\mathcal{FP}$ of the employed fusion operators with their respective equations.}
\vspace{1ex}
\fontsize{15}{15}\selectfont
\scalebox{.58}{
\label{tab:fusion_ops}
  \begin{tabular}{|c|c|}
    \hline
    \textbf{Fusion Operator} & \textbf{Mathematical Formula} \\
    \hline
    $\mathrm{Sum}$  & $ \mathrm{Sum}(X,Y) = X+Y $ \\
    \hline
     $\mathrm{Attention}$ & $\mathrm{ScaleDotAttn}(X,Y) = \mathrm{Softmax}(\frac{X . Y^T}{\sqrt{C}}).Y$ \\
    \hline
     $\mathrm{LinearGLU}$  & $\mathrm{LinearGLU}(X, Y) = \mathrm{GLU}(X W_1, Y W_2) = X W_1 \odot \mathrm{Sigmoid}(Y W_2) $ \\
    \hline
     $\mathrm{ConcatFC}$ & $ \mathrm{ConcatFC}(X,Y) =  \mathrm{ReLU} (\mathrm{Concat}(X,Y) . W + b) $ \\
    \hline
     $\mathrm{Squeeze-Excitation}$ &  {$\mathrm{SE}(X,Y) =  E_X \odot Y$ $|$ $E_X = \sigma(S_X . W + b) |$  $S_X = \frac{1}{L}\sum_{i=1}^{L} X(B,C,i) $}  \\
    \hline
     $\mathrm{ConcatMish}$ & $\mathrm{ConcatMish}(X, Y) = \mathrm{Mish}(\mathrm{Cat}(X W_1, Y W_2))$ $|$ $\mathrm{Mish}(Z)=\tanh(\log(1 + \exp(X)) \odot \mathrm{Sigmoid}(Y W_2)$  \\
    \hline
  \end{tabular}
}
\end{table}
\vspace{-2ex}
\begin{enumerate}
    \item $\mathrm{Sum}$: In the context of multimodal fusion, there are instances where cross-modality interactions exhibit an additive nature. In scenarios where the modalities are relatively independent, aggregating their representations through summation offers a straightforward means of capturing joint information that incorporates each modality's distinctive strengths. 
   
    \item $\mathrm{ScaleDotAttn}$: Motivated by the promising outcomes of the Attention mechanism in modeling cross-modality interactions \cite{nagrani2021attention}, we incorporate the scaled dot-product attention as a fusion operator to investigate its efficiency in addressing multimodal tasks.

    \item $\mathrm{LinearGLU}$: Through performing element-wise multiplication $\odot$ on the linearly transformed modality $X$ and the sigmoid-gated of modality $Y$, this operator allows modality $Y$ to determine the contribution of each element from the modality $X$.
    
    \item $\mathrm{ConcatFC}$: Here the unimodal features are concatenated on the channel dimension $C$. Then a linear transformation is applied and followed by the $ReLU$ activation.

    \item $\mathrm{Squeeeze-Excitation}$: The utilization of the Squeeze-and-Excitation module \cite{hu2018squeeze} for channel-wise recalibration has demonstrated its effectiveness when applied on various blocks of the unimodal convolutional neural network (CNN). We further extend its applicability to the multimodal case and include it as a fusion operator.

    \item $\mathrm{CatConvMish}$ : This operator provides a fusion mechanism that captures both linear and non-linear interactions between the modalities through a series of primitive operations.
    
\end{enumerate}

\subsubsection{Differentiable search strategy}

In section \ref{fusion_network_search_space}, we defined the search space for the fusion network. This search space also includes parameters related to the fusion macro-architecture, such as $\mathcal{C}$, the number of fusion cells, and $\mathcal{D}$, the number of fusion nodes within each cell. We note that these parameters are also searchable within the first evolutionary search stage, ensuring a diverse range of fusion macro-architectures. While in the \textit{second-stage}, we search for the fusion micro-architecture using DARTS for a priori sampled combination of $\mathcal{C}$ and $\mathcal{D}$ in the \textit{first-stage} of \textit{Harmonic-NAS}.

During the \textit{second-stage}, the weights $\alpha$, $\beta$, and $\gamma$ are jointly updated. This involves the use of gradient-based optimization, which allows for the exploration of various fusion micro-architecture configurations by training a hypernetwork with the following loss function $\mathcal{L}_{fusion}$:
\begin{equation*}
    \min_{\alpha, \beta, \gamma} \mathcal{L}_{fusion}(\mathcal{H}, Z, \alpha, \beta, \gamma, {\mathcal{Y}}) 
\end{equation*}
\begin{equation*}
    \mathcal{L}_{fusion}(\mathcal{H}, Z, \alpha, \beta, \gamma, {\mathcal{Y}}) \ = \bigr[\mathcal{L}_{task}(\mathcal{H}, Z, \alpha, \beta, \gamma, {\mathcal{Y}})\bigr]^{\ a} \ + \ \bigr[\mathcal{L}_{Lat}(\mathcal{H}, \gamma)\bigr]^{\ b}
    + \ \bigr[\mathcal{L}_{Enrg}(\mathcal{H}, \gamma)\bigr]^{\ c}
\end{equation*}
where $\mathcal{H}$ is the initial fusion hypernet that represents all the possible configurations for DARTS, 
$\gamma$ are the weights that control the fusion operator selection within the fusion cells, $Z$ denotes the multimodal input data, and $\mathcal{Y}$ are the ground truth labels. The exponents $a$, $b$, and $c$ serve as control knobs for the importance of each performance metric in the overall loss function $\mathcal{L}_{fusion}$. By adjusting these exponents, we can emphasize particular objectives such as task performance, latency, or energy consumption during the search process of the fusion network. The task loss $\mathcal{L}_{task}$ varies depending on the target task (e.g., cross-entropy, binary cross-entropy). The hardware loss $\mathcal{L}_{(Lat||Enrg)}$ involves hardware-specific metrics -- It takes as input the architectural specification of the fusion network and a lookup table ($LUT_{device}$) for latency and energy measurements of the fusion operators on the target device. The hardware loss is then computed as follows:
\begin{equation}
    \begin{split}
        \mathcal{L}_{(Lat||Enrg)}(\mathcal{H}, \gamma, LUT_{device})  = \sum\limits_{p=1}^{\mathcal{C}} \sum\limits_{d=1}^{\mathcal{D}} \gamma_{p} [d] \ . \ LUT_{device}(\mathcal{FP}, (Lat||Enrg))
    \end{split}
\end{equation}
We also note that we first apply the \textit{softmax} operation on the $\gamma$ weights so that $\gamma_p[d]$ represents the probabilities of each fusion operator to be selected for the $d^{th}$ node within the $p^{th}$ cell, where $LUT_{device}$ contains hardware measurements for each fusion operator within $\mathcal{FP}$. 

%% file: Sections/evaluation.tex
\section{Experiments}\label{sec:eval}
\subsection{Experimental Setup}
\subsubsection{Multimodal tasks and datasets}
To evaluate the effectiveness of \textit{Harmonic-NAS} in designing efficient multimodal networks, we conduct experiments on various multimodal datasets as listed in Table \ref{tab:mm_datasets}.
\vspace{-0.5cm}
\begin{table}[ht!]
\centering
\caption{Multimodal Datasets and tasks used by \textit{Harmonic-NAS}}
\vspace{1ex}
\fontsize{15}{15}\selectfont
\scalebox{.65}{
\label{tab:mm_datasets}
\begin{tabular}{|c|c|c|c|}
\hline
\textbf{Dataset} & \textbf{Modalities} & \textbf{Samples (train; val; test)} & \textbf{Task} \\
\hline
AV-MNIST & Image, Audio & \{55000; 10000; 5000\} & Digit classification \\
\hline
MM-IMDB & Image, Text & \{15552; 2608; 7799\} & Movie genres classification\\
\hline
HARM-P & Image, Text & \{3020; 177; 355\} & Harmful US-Politics-related memes detection \\
\hline
\end{tabular}
}
\end{table}

\noindent{\large \textcircled{\small 1}} \textbf{AV-MNIST:} The audio-visual MNIST for hand handwritten digits classification. It includes two modalities: image samples of handwritten digits and audio samples of spoken digits.

\noindent{\large \textcircled{\small 2}} \textbf{MM-IMDB:} The multi-modal IMDB dataset for multi-label classification of movie genres using movie titles and metadata as textual modality as well as movie posters as a visual modality.

\noindent{\large \textcircled{\small 3}} \textbf{Harmful Memes:} We use the \textit{Harm-P} dataset \cite{pramanick2021momenta} for detecting harmful memes related to United States politics. The dataset contains memes images collected from various social media platforms and text extracted from the meme image using Google’s OCR Vision API\footnote{\url{cloud.google.com/vision/docs/ocr}}.

\subsubsection{Unimodal Backbones Settings}
We build our unimodal supernets for image and audio processing upon the once-for-all framework \cite{cai2019once}. We further adjust their original search space by reducing the number of neural blocks to 3 and 5 for the \textit{AV-MNIST} and \textit{Memes-P} datasets, respectively. 
For each block: The depth is chosen from $\{2, 3, 4\}$, the width expansion ratio for each layer within the block is selected from $\{3, 4, 6\}$, and the kernel size is picked from $\{3, 5, 7\}$. Overall, the search space complexity ranges between $\mathcal{O}$(2$\times$10$^{5}$) and $\mathcal{O}$(2$\times$10$^{7}$). For textual modality, we use a Maxout network to process the text embedding on MM-IMDB dataset and HARM-P.
To train our backbones, we use Adam as an optimizer with weight decay of $1e^{-4}$ and a cosine annealing as a learning rate scheduler with a base learning rate of $1e^{-3}$. An early-stopping is used to determine the number of training epochs. 

\subsubsection{Hardware Experimental Settings}
The multimodal workloads are deployed using the Pytorch 1.12 framework running on top of CUDA 11.4 and cuDNN 8.3.2. We target the following Edge devices provided by NVIDIA: \textbf{(i)} Jetson AGX Xavier equipped with an NVIDIA Carmel Arm-64bit CPU and a high-performance Volta GPU of 512 GPU cores and 64 Tensor cores. \textbf{(ii)} Jetson TX2 composed of an NVIDIA Denver 64Bit and ARM-A57 CPU cores along a high-performance Pascal GPU of 256 GPU cores.

\subsubsection{Evolutionary and Differentiable Search Settings}
In our optimization process, we run the evolutionary search for 30 generations, each with a population size of 128 individuals.
The top quartile of the promising backbones is selected for the fusion search to derive multimodal networks. Then, half of the elite multimodal networks are chosen for mutation and crossover on their unimodal backbones to generate the next population of the evolutionary search.
The mutation and crossover probabilities are set to 0.4 and 0.8, respectively. We run the differentiable fusion search using DARTS for 25 epochs using Adam as an optimizer and a cosine-annealing learning rate scheduler. The learning rate ranged from $e^{-6}$ to $e^{-4}$.

\begin{figure}[ht!]
\centering
    \includegraphics[width=1\textwidth]{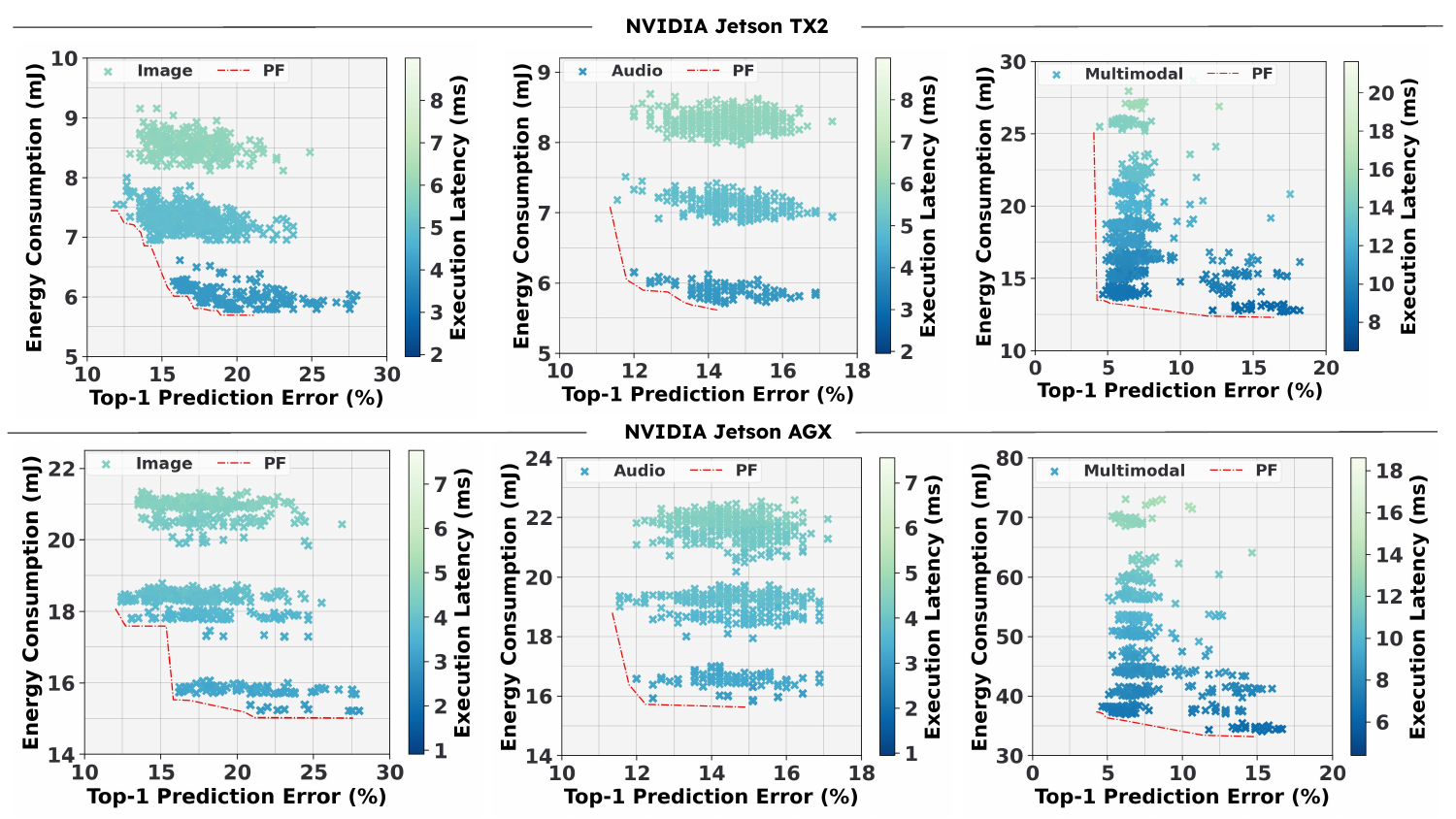}
    \caption{The first two columns (left to right) show the explored unimodal backbones in the first search stage for image and audio modalities, respectively. The last column shows the explored multimodal networks in the second search stage. The first and second rows report results on the NVIDIA TX2 and AGX devices, respectively. The red Pareto Front (PF) highlights the models with the best tradeoff between performance metrics.} 
    \label{fig:avmnist_results}
\end{figure}

\subsection{Experimental Results}
\subsubsection{Two-tier optimization results analysis}
To showcase the efficiency of the two-tier optimization of \textit{Harmonic-NAS}, we report the search results for the AV-MNIST dataset on two hardware devices in Figure \ref{fig:avmnist_results}. The two first columns depict the explored unimodal backbones in the first search stage for image and audio, respectively, whereas the last column shows the explored multimodal networks in the second search stage.
From the reported result, the explored unimodal backbones in the first search stage exhibit high variation in TOP-1 error ranges from $\sim$\textbf{12\%} to $\sim$\textbf{27\%} for image modality and from $\sim$\textbf{12\%} to $\sim$\textbf{18\%} for audio modality. By searching for optimal fusion networks for the two modalities in the second search stage, as shown in the third column, the TOP-1 error ranges from $\sim$\textbf{4\%} to $\sim$\textbf{17\%}, improving further the performance of the unimodal backbones by up to \textbf{63\%} error decrease. We can also notice the search intensification in the region that provides low TOP-1 errors in the multimodal case where more than \textbf{$\sim$60\%} of explored networks exhibit a TOP-1 error less than $\sim$\textbf{8\%}. 
On the hardware efficiency, latency and energy values are generally doubled for the multimodal case as backbones and fusion are executed in a sequential pipeline manner. However, by integrating the hardware loss, \textit{Harmonic-NAS} was able to identify a diverse set of optimal solutions in the Pareto front with a good compromise between prediction TOP-1 error, latency, and energy consumption. 

\subsubsection{Pareto optimal multimodal models analysis}

In this section, we provide an in-depth analysis of the Pareto optimal backbone and multimodal models obtained by \textit{Harmonic-NAS} and compare them against SoTA multimodal-NAS approaches on different Edge devices. In the following, we analyze the results of each multimodal dataset.

\begin{wraptable}{r}{0.6\textwidth}
\centering
\vspace{-5mm}
\caption{Performance evaluation on AV-MNIST dataset}
\vspace{1ex}
\label{tab:pf_avmnist}
\fontsize{9}{9}\selectfont
\scalebox{0.7}
{
\begin{tabular}{lcccc}
\hline
\textbf{SoTA work}&
  \textbf{Modality}&
  \textbf{Acc(\%)}&
  \textbf{Latency(ms)}&
  \textbf{Energy(mJ)}\\ \hline
\multicolumn{5}{c}{Unimodal Backbones} \\ \hline
MVAE \cite{wu2018multimodal} &
  Image &
  65.10 &
 \textbf{A}:1.35,\textbf{T}:2.62 &
 \textbf{A}:4.82,\textbf{T}:2.71 \\
MFAS \cite{perez2019mfas} &
  Image &
  74.52 &
 \textbf{A}:1.03,\textbf{T}:1.93 &
 \textbf{A}:3.78,\textbf{T}:2.11 \\
\text{Harmonic-NAS} (\textbf{T}:TX2) &
  Image &
 \textbf{88.00} &
  4.97 &
 \textbf{7.54} \\
\text{Harmonic-NAS} (\textbf{A}:AGX) &
  Image &
 \textbf{87.55} &
 \textbf{3.91} &
  18.26 \\ \hline
MVAE \cite{wu2018multimodal} &
  Audio &
  42.00 &
 \textbf{A}:1.92,\textbf{T}:4.12 &
 \textbf{A}:9.73,\textbf{T}:11.71 \\
MFAS \cite{perez2019mfas} &
  Audio &
  66.06 &
 \textbf{A}:1.78,\textbf{T}:2.92 &
 \textbf{A}:6.98,\textbf{T}:5.50 \\
\text{Harmonic-NAS} (\textbf{T}:TX2) &
  Audio &
 \textbf{88.44} &
  4.97 &
 \textbf{7.18} \\
\text{Harmonic-NAS} (\textbf{A}:AGX) &
  Audio &
 \textbf{88.44} &
 \textbf{3.74} &
  18.99 \\ \hline
\multicolumn{5}{c}{Multimodal Neural Networks} \\ \hline
MVAE \cite{wu2018multimodal} &
  Image + Audio &
  72.30 &
 \textbf{A}:4.40,\textbf{T}:8.53 &
 \textbf{A}:15.64,\textbf{T}:15.67 \\
MFAS \cite{perez2019mfas} &
  Image + Audio &
  88.38 &
 \textbf{A}:4.37,\textbf{T}:6.41 &
 \textbf{A}:10.76,\textbf{T}:10.67 \\
BM-NAS \cite{yin2022bm} &
  Image + Audio &
 \textbf{91.11} &
 \textbf{A}:3.26,\textbf{T}:5.41 &
 \textbf{A}:14.17,\textbf{T}:8.60 \\ \hline
\multirow{3}{*}{\textbf{\text{Harmonic-NAS} (TX2)}} &
  \multirow{3}{*}{Image + Audio} &
  92.88 &
 \textbf{8.96} &
  13.93 \\
 &
   &
 \textbf{95.55} &
  14.41 &
  25.49 \\
 &
   &
  95.33 &
  9.11 &
  \textbf{13.88} \\ \hline
\multirow{3}{*}{\textbf{\text{Harmonic-NAS} (AGX)}} &
  \multirow{3}{*}{Image + Audio} &
  94.22 &
 \textbf{6.95} &
  \textbf{37.17} \\
 &
   &
 \textbf{95.33} &
  7.26 &
  38.24 \\
 &
   &
  95.11 &
  7.19 &
  38.14 \\ \hline
\end{tabular}
}
\vspace{-4mm}
\end{wraptable}

\noindent{\large \textcircled{\small 1}} \textbf{The Audio-Visual MNIST:}
Table \ref{tab:pf_avmnist} reports the obtained performance metrics (Top-1 Accuracy, latency, and energy) of the Pareto optimal backbones and multimodal models found by \textit{Harmonic-NAS} compared against SoTA counterparts. In the rest of the paper, for abbreviation, latency and energy values on the Jetson AGX and TX2 are preceded by letters \textbf{'A'} and \textbf{'T'}, respectively.
Firstly, we noticed that the use of LeNet-based backbones in SoTA models such as \cite{wu2018multimodal} and \cite{perez2019mfas} limits the unimodal feature extraction capability even if it's boosted with a powerful fusion architecture search strategy as in BM-NAS \cite{yin2022bm}. 
Thus, the low-performing unimodal backbones limit the multimodal performances. However, by optimizing the unimodal backbones, an accuracy improvement of up to $\sim$\textbf{4.44\%} has been obtained over the best SoTA multimodal model while enjoying low latency and energy levels. This is attributed to the hierarchical design space that first searches for an optimal network to learn the unimodal embedding of features and then proceeds to optimize the joint embedding through the fusion architecture search.
Furthermore, \textit{Harmonic-NAS} has also shown an adaptation to different hardware devices with varying computational capacities. The Pareto fronts in Figure \ref{fig:avmnist_results} and the selected Pareto models in Table \ref{tab:pf_avmnist} show such hardware efficiency diversity, highlighting further the effectiveness of \textit{\text{Harmonic-NAS}} in designing efficient MM-NNs on Edge devices.

\noindent{\large \textcircled{\small 2}} \textbf{The Multimodal IMDB:}
In Table \ref{tab:pf_imdb}, we show the performance of the Pareto optimal backbones and multimodal models of \textit{\text{Harmonic-NAS}} compared to SoTA methods on MM-IMDB dataset.
\begin{wraptable}{r}{0.6\textwidth}
\centering
\vspace{-5mm}
\caption{Performance evaluation on MM-IMDB dataset}
\vspace{1ex}
\label{tab:pf_imdb}
\fontsize{9}{9}\selectfont
\scalebox{0.67}
{
\begin{tabular}{lcccc}
\hline
\textbf{SoTA work}                           & \textbf{Modality}             & \textbf{F1-W(\%)} & \textbf{Latency(ms)} & \textbf{Energy(mJ)} \\ \hline
\multicolumn{5}{c}{Unimodal Backbones}                                                                                                        \\ \hline
Maxout \cite{yin2022bm}                                   & Text                          & 57.54             & \textbf{A}:0.82, \textbf{T}:0.99       & \textbf{A}:3.33, \textbf{T}:1.27      \\
Maxout \textbf{(Ours)}                                & Text                          & \textbf{61.18}    & \textbf{A}:0.62, \textbf{T}:1.09       & \textbf{A}:2.87, \textbf{T}:1.40      \\ \hline
VGG19                                        & Image                         & 49.21             & \textbf{A}:30.91, \textbf{T}:117.94    & \textbf{A}:791.16, \textbf{T}:1013.09 \\
\text{Harmonic-NAS} (\textbf{T}:TX2)                           & Image                         & \textbf{46.12}    & 39.82                & \textbf{248.05}     \\
\text{Harmonic-NAS} (\textbf{A}:AGX)                           & Image                         & \textbf{46.12}    & \textbf{20.09}       & 264.34              \\ \hline
\multicolumn{5}{c}{Multimodal Neural Networks}                                                                                                \\ \hline
MFAS \cite{perez2019mfas}                                        & Image + Text                  & 62.50             & \textbf{A}:32.77, \textbf{T}:119.97    & \textbf{A}:800.2, \textbf{T}: 1016.4  \\
BM-NAS \cite{yin2022bm}                                      & Image + Text                  & \textbf{62.92}    & \textbf{A}:32.78, \textbf{T}:119.97    & \textbf{A}:800.66, \textbf{T}:1016.58 \\ \hline
\multirow{3}{*}{\textbf{\text{Harmonic-NAS} (TX2)}} & \multirow{3}{*}{Image + Text} & 63.61             & \textbf{21.37}       & \textbf{113.99}     \\
                                             &                               & \textbf{64.36}    & 28.68                & 163.04              \\
                                             &                               & \textbf{64.27}    & 23.67                & 121.75              \\ \hline
\multirow{3}{*}{\textbf{\text{Harmonic-NAS} (AGX)}} & \multirow{3}{*}{Image + Text} & 63.75             & \textbf{11.32}       & \textbf{130.42}     \\
                                             &                               & \textbf{64.36}    & 14.29                & 177.55              \\
                                             &                               & \textbf{64.27}    & 13.05                & 140.00              \\ \hline
\end{tabular}
}
\vspace{-4mm}
\end{wraptable}
As a metric for the prediction performance, we use the weighted F1-score (F1-W) instead of the TOP-1 accuracy to account for the imbalanced data distribution in the MM-IMDB dataset. As shown, \textit{Harmonic-NAS} achieves superior results, surpassing the best results achieved by SoTA works such as BM-NAS \cite{yin2022bm} with up to $\sim$\textbf{1.45\%} improvement in the weighted F1-score. On the hardware side, \textit{Harmonic-NAS} models are $\sim$\textbf{4.18x} and $\sim$\textbf{2.19x} more latency and energy efficient compared to BM-NAS multimodal \cite{yin2022bm} models.
Additionally, the least accurate multimodal network in \textit{Harmonic-NAS} is $\sim$\textbf{2.89x} and $\sim$\textbf{6.14x} more latency and energy efficient than the most accurate SoTA model on the Jetson AGX. This further demonstrates our framework's superiority in adapting to scenarios where hardware efficiency is prioritized.

\noindent{\large \textcircled{\small 3}} \textbf{The Harmful Politics Memes:}
In Table \ref{tab:pf_memes}, we report the performance of the Pareto optimal backbones and multimodal models of \textit{Harmonic-NAS} and SoTA works on Memes-P dataset.
\begin{wraptable}{r}{0.6\textwidth}
\centering
\vspace{-5mm}
\caption{Performance evaluation on Memes-P dataset}
\vspace{1ex}
\label{tab:pf_memes}
\fontsize{9}{9}\selectfont
\scalebox{0.65}
{
\begin{tabular}{lcccc}
\hline
\multicolumn{1}{c}{\textbf{SoTA work}}       & \textbf{Modality}             & \textbf{Acc(\%)} & \textbf{Latency(ms)} & \textbf{Energy(mJ)}  \\ \hline
\multicolumn{5}{c}{Unimodal Backbones}                                                       \\ \hline
TextBERT           & Text         & 74,55          & \textbf{A}:22.71, \textbf{T}:59.63  & \textbf{A}:524,28, \textbf{T}:461.91  \\
Maxout \textbf{(Ours)}      & Text         & \textbf{83.38} & \textbf{A}:0.61, \textbf{T}:1.08    & \textbf{A}:2.82, \textbf{T}:1.21      \\ \hline
VGG19              & Image        & 73.65          & \textbf{A}:31.01, \textbf{T}:116.80 & \textbf{A}:786.94, \textbf{T}:1001.17 \\
DenseNet-161       & Image        & 71.80          & \textbf{A}:28.59, \textbf{T}:92.32  & \textbf{A}:589.54, \textbf{T}:710.21  \\
ResNet-152         & Image        & 71.02          & \textbf{A}:29.80, \textbf{T}:92.58  & \textbf{A}:754.56, \textbf{T}:798.52  \\
Harmonic-NAS (\textbf{T}:TX2) & Image        & \textbf{84.78} & 10.40             & \textbf{28.88}      \\
Harmonic-NAS (\textbf{A}:AGX) & Image        & \textbf{84.78} & \textbf{5.68}     & 38.28               \\ \hline
\multicolumn{5}{c}{Multimodal Neural Networks}                                               \\ \hline
ViLBERT CC \cite{lu2019vilbert}        & Image + Text & 84.66          & \textbf{A}:22.58, \textbf{T}:59.93  & \textbf{A}:523.93, \textbf{T}:449.89  \\
MOMENTA \cite{pramanick2021momenta}                                      & Image + Text                  & \textbf{87.14}   & \textbf{A}:35,09, \textbf{T}:176,43    & \textbf{A}:1311,22, \textbf{T}:1463.08 \\ \hline
\multirow{3}{*}{\textbf{Harmonic-NAS (TX2)}} & \multirow{3}{*}{Image + Text} & 88.45            & \textbf{10.51}       & 25.63                \\
                   &              & \textbf{90.42} & 12.47             & 31.92               \\
                   &              & \textbf{90.14} & 11.11             & 26.63               \\ \hline
\multirow{3}{*}{\textbf{Harmonic-NAS (AGX)}} & \multirow{3}{*}{Image + Text} & 88.16            & \textbf{5.79}        & 38.34                \\
                   &              & \textbf{90.42} & 7.51              & 50.79               \\
                   &              & \textbf{90.14} & 6.91              & 43.49               \\ \hline
\end{tabular}
}
\vspace{-4mm}
\end{wraptable}
Compared to SoTA works, our framework provides better unimodal and multimodal neural networks regarding accuracy and hardware efficiency. The optimal multimodal networks found by \textit{Harmonic-NAS} for both hardware devices have shown an improvement of up to $\sim$\textbf{3\%} in the TOP-1 accuracy while being $\sim$\textbf{6.06x} more latency efficient than MOMENTA \cite{pramanick2021momenta} on the AGX. In light of highlighting the hardware awareness of \textit{Harmonic-NAS}, we also found another MM-NN with a slight drop in accuracy but more suitable for resource-constrained scenarios, exhibiting $\sim$\textbf{1.19x} and 
$\sim$\textbf{1.32x} gains in latency and energy, respectively, compared to the most accurate MM-NN from \textit{Harmonic-NAS}.

\subsubsection{On the importance of the hierarchical design for multimodal networks}
\begin{wrapfigure}{r}{0.75\textwidth}
  \centering
  \vspace{-4mm}
  \hspace{-2mm}
    \includegraphics[width=0.75\textwidth]{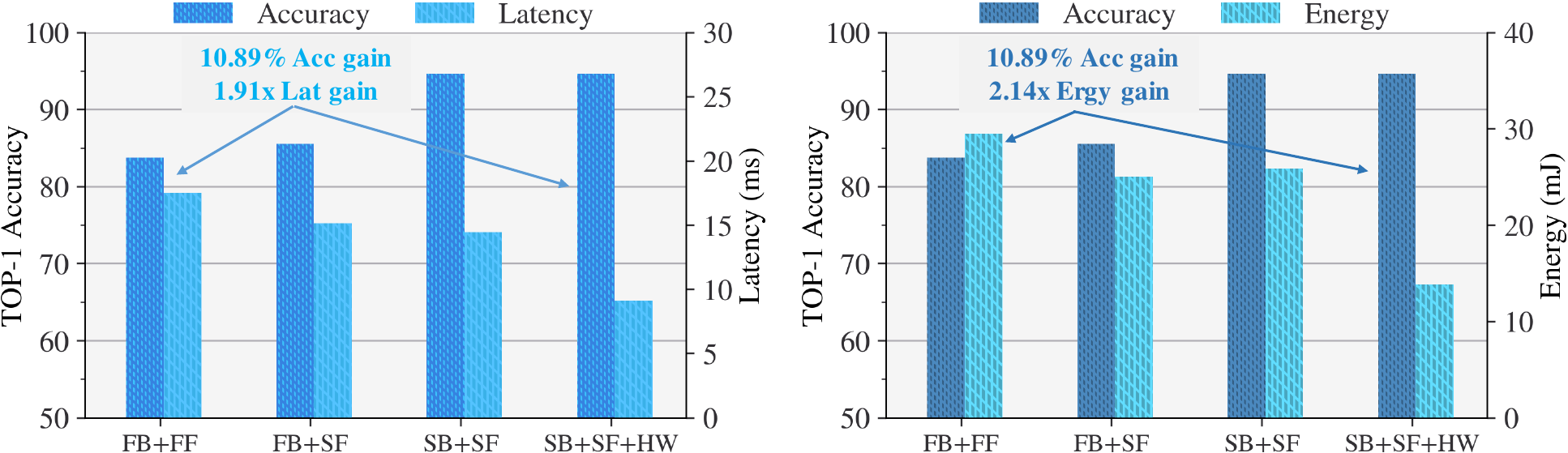}
  \vspace{-3.5mm}
  \caption{Performance of the progressive design of multimodal networks}
  \vspace{-3.5mm}
  \label{fig:progressive}
\end{wrapfigure}

To further demonstrate the importance of the considered neural design parameters for multimodal networks in \textit{Harmonic-NAS}, we provide Figure \ref{fig:progressive} in which we report the obtained performances when progressively adding new design dimensions to the search space. In the x-axis, we refer to the fixed and searched backbones by \textbf{'FB'} and \textbf{'SB'}, respectively. The fixed and searchable fusion networks are refereed by \textbf{'FF'} and \textbf{'SF'}, respectively. The term \textbf{'HW'} in the x-axis designates the inclusion of the Hardware efficiency metrics (i.e., latency and energy) as optimization objectives.
In the first case \textit{(FB+FF)}, the unimodal backbones are fixed for each modality to the subnets with the highest learning capacity from our supernets (i.e., the subnets with the maximum number of learnable weights), whereas the fusion is selected as a dense architecture with all the fusion possibilities. In the second and third cases, we activate the search for the fusion network in \textit{(FB+SF)} and unimodal backbones in \textit{(SB+SF)}. All the results in Figure \ref{fig:progressive} are reported for the AV-MNIST dataset on the NVIDIA Jetson TX2.
From comparing \textit{(FB+FF)} and \textit{(FB+SF)}, we notice that simply optimizing the fusion network for fixed backbones does not always result in optimal multimodal models. This supports our assumption of the importance of jointly optimizing the unimodal and multimodal feature embedding, as the best standalone unimodal backbones are not initially designed nor trained for multimodal learning.
In \textit{(SB+SF)}, incorporating the unimodal backbones optimization yields an accuracy improvement of $\sim$\textbf{11\%}. This is because our first-stage search engine could identify unimodal backbones tailored for the multimodal scenario, even if their unimodal performances are lower than that of the maximum subnets. Furthermore, by adding the Hardware metrics as optimization objectives in the \textit{(SB+SF+HW)} case, latency and energy gains of $\sim$\textbf{1.91x} and $\sim$\textbf{2.14x}, respectively, have been obtained compared to the \textit{(FB+FF)} case while ensuring the same accuracy level, emphasizing further the importance of considering the hardware efficiency when designing multimodal networks on resource-constrained Edge devices.

%% file: Sections/conclusion.tex
\section{Conclusion}\label{sec:conclusion}
In this paper, we presented \textit{Harmonic-NAS}, a novel framework for the joint optimization of unimodal backbones and fusion networks to learn an effective joint embedding of features from multiple modalities. \textit{Harmonic-NAS} employs a two-tier optimization scheme with an evolutionary search stage for the unimodal backbone networks and a differentiable search stage for the fusion architecture design. \textit{Harmonic-NAS} also includes the hardware dimension in its optimization procedure by integrating metrics such as latency and energy consumption. Evaluation results have seen the superiority of \textit{Harmonic-NAS} over SoTA multimodal-NAS approaches in discovering efficient multimodal networks with up to $\sim$\textbf{10.9\%} accuracy improvement, $\sim$\textbf{1.91x} latency reduction, and $\sim$\textbf{2.14x} energy efficiency gain on different Edge devices.

%% file: Sections/appendix.tex
\appendix

\section{The Fusion Operator Set Design }\label{apd:first}

As a baseline, we considered the fusion search space of recent works MFAS ($\mathrm{ConcatFC}$) and BM-NAS ($\mathrm{Attention}$, $\mathrm{LinearGLU}$, $\mathrm{Sum}$) to design our initial $\mathcal{FP}$ set. We conducted a preliminary analysis and included other fusion operators, notably, $\mathrm{ConcatMish}$ and $\mathrm{Squeeze-Excitation}$, yielding better accuracy and efficiency tradeoffs. Consequently, these new added fusion operators have also contributed to the superior results of our multimodal models over SoTA methods.

\begin{table}[ht!]
\centering
\caption{Backbones configurations of one of our optimal MM-NNs on the AV-MNIST dataset.} 
\label{tab:abl_back}
\fontsize{9}{9}\selectfont
\scalebox{1.2}{
\begin{tabular}{ccccccc} 
\hline
\textbf{Modality} & \textbf{Kernel size} & \textbf{Expand ratio} & \textbf{Depth} & \textbf{Acc (\%)} & \textbf{Lat (ms)} & \textbf{Ergy (mJ)}  \\ 
\hline
Image             & {[}5, 5, 5, 7]       & {[}3, 6, 4, 3]        & 2              & 82.66              & 4.02             & 6.01             \\ 
\hline
Audio             & {[}3, 3, 7, 5]       & {[}3, 3, 3, 6]        & 2              & 85.55              & 3.95             & 5.71             \\
\hline
\end{tabular}
}
\end{table}

\noindent To better understand the impact of the fusion operator, we examine one of our optimal MM-NN models on the AV-MNIST dataset (Table \ref{tab:pf_avmnist}, TX2, Acc=95.33\%, Lat=9.11ms, Ergy=13.88mJ). The backbones configurations for image and audio modalities are summarized in Table \ref{tab:abl_back}.

\begin{table}[ht!]
\centering
\caption{The impact of the fusion operator on the MM-NN performance on AV-MNIST.} 
\label{tab:abl_back_fuse_avmnist}
\fontsize{9}{9}\selectfont
\scalebox{1.2}{
\begin{tabular}{lccc} 
\hline
\textbf{Fusion operator}                                                                            & \textbf{Acc (\%)} & \textbf{Lat (ms)} & \textbf{Ergy (mJ)}  \\ 
\hline
\begin{tabular}[c]{@{}l@{}}\textbf{*Searchable (Ours)}\\\textbf{(ConcatMish, ConcatFC)}\end{tabular} & \textbf{95.33}          & \textbf{9.11}     & \textbf{13.88}     \\ 
\hline
Sum                                                                                                 & 93.70                    & 8.09              & 12.32              \\
Attention                                                                                           & 89.55                   & 8.98              & 13.13              \\
LinearGLU                                                                                           & 94.22                   & 9.03              & 14.14              \\
ConcatFC                                                                                            & 94.66                   & 9.02              & 13.77              \\
\hline
\end{tabular}
}
\end{table}

\noindent In this ablation study, we maintain the unimodal backbones and optimal found fusion macro-architecture (i.e., number of fusion cells and nodes) by our \textit{first-stage} optimization engine and only vary the fusion operators. The results are reported for the AV-MNIST dataset in Table \ref{tab:abl_back_fuse_avmnist}. As shown, the contradictory nature of objectives is explicit as more accuracy yields high latency and energy. Notably, our newly added fusion operator, $\mathrm{ConcatMish}$ with the existing $\mathrm{ConcatFC}$, depict the optimal trade-off. \\

\begin{longtblr}[
  label = tab:best_meme_config,
  entry = {Backbones configurations of one of our optimal MM-NNs on the Memes-P dataset.},
  caption = {Backbones configurations of one of our optimal MM-NNs on the Memes-P dataset.},
]{
  width = \linewidth,
  colspec = {Q[85]Q[250]Q[250]Q[85]Q[70]Q[90]Q[83]},
  cells = {c},
  cell{1}{2} = {c=3}{0.585\linewidth},
  cell{2}{2} = {c=3}{0.585\linewidth},
  hline{1,2-4,5} = {-}{},
}
\textbf{Modality} & \textbf{Configuration}                                                  &                             &                & \textbf{Acc (\%)} & \textbf{Lat (ms)} & \textbf{Ergy (mJ)} \\
Text              & Maxout=\{hidden\_features: 128, n\_blocks: 2, factor\_multiplier: 2\} &                             &                &   83.38            &   1.08               &   1.21              \\
                  & \textbf{Kernel size}                                                    & \textbf{Expand ratio}       & \textbf{Depth} &               &                  &                 \\
Image             & {[}3,3,3,3,5,3,3,3,5,5,3,5]                                             & {[}4,3,4,6,4,4,3,6,6,6,3,4] & {[}2,3,2]      & 85.91         & 10.94            & 25.44          
\end{longtblr}

\noindent Similarly, on the Memes-P dataset, we conducted a the same analysis on one of our optimal MM-NN (See Table \ref{tab:pf_memes}, TX2, Acc=90.42\%, Lat=12.47ms, Ergy=31.92mJ). The backbones configurations for text and image modalities are summarized in Table \ref{tab:best_meme_config}.

\begin{table}[ht!]
\centering
\caption{The impact of the fusion operator on the MM-NN performance on Memes-P} 
\label{tab:abl_back_fuse_memes}
\fontsize{9}{9}\selectfont
\scalebox{1.2}{
\begin{tabular}{lccc} 
\hline
\textbf{Fusion operator}                                                                               & \textbf{Acc (\%)} & \textbf{Lat (ms)} & \textbf{Ergy (mJ)}  \\ 
\hline
\begin{tabular}[c]{@{}l@{}}\textbf{*Searchable (Ours)}\\\textbf{(Sum,Squeeze-Excitation)}\end{tabular} & \textbf{90.42}    & \textbf{12.47}    & \textbf{31.92}     \\ 
\hline
Sum                                                                                                    & 88.73             & 12.38             & 28.44              \\
Attention                                                                                              & 89.01             & 15.04             & 30.86              \\
LinearGLU                                                                                              & 89.29             & 15.18             & 33.89              \\
ConcatFC & 89.29 & 15.16 & 32.78 \\
\hline
\end{tabular}
}
\end{table}

As shown in Table The newly added $\mathrm{Squeeze-Excitation}$ operator with the existing $\mathrm{Sum}$ yield better results and balance between accuracy, latency, and energy (See Table \ref{tab:abl_back_fuse_memes}). Thus further demonstrating the fusion operators' diversity across different tasks, modalities, and datasets.

\section{Visualizations of our learned MM-NNs}\label{apd:second}
In the following, we provide visualizations of the learned fusion architectures on various multimodal datasets. We note that our MM-NN models are built upon different backbones that technically share the same macro-architecture -as fixed by the OFA supernet design-. However, as our unimodal backbones are searchable, the inner structure of the neural blocks is different from one MM-NN to another. We refer the reader to Tables \ref{tab:pf_avmnist}, \ref{tab:pf_imdb}, and \ref{tab:pf_memes} for more details on the unimodal backbones performance for each reported multimodal representation. The following MM-NN visualizations are all reported for the NVIDIA Jetson TX2 device.

\begin{figure}[h]
\centering
    \includegraphics[width=0.8\textwidth]{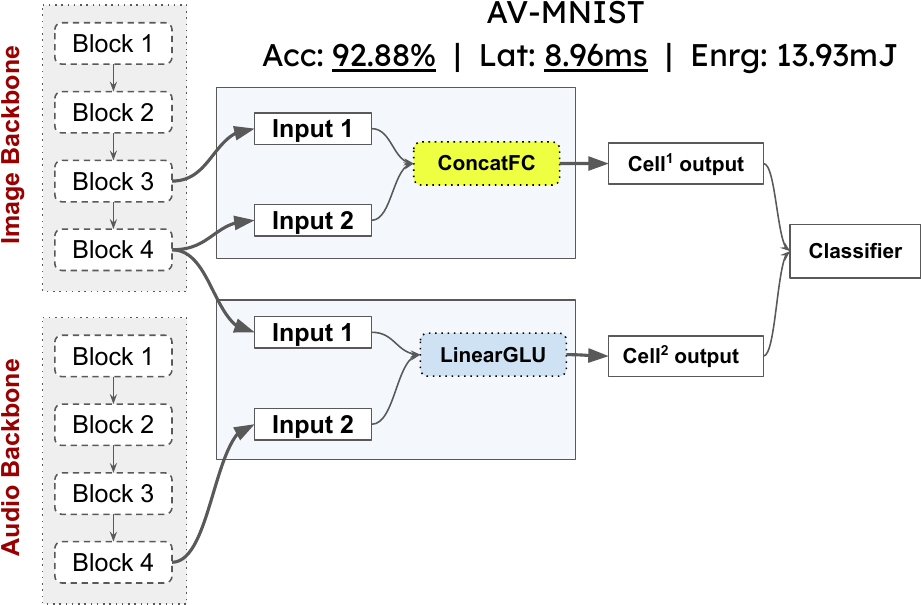}
    \caption{Visualization of the latency-efficient MM-NN for the AV-MNIST dataset on the TX2.}
    \label{fig:avmnist_1}
\end{figure}

\begin{figure}[tp]
\centering
    \includegraphics[width=0.9\textwidth]{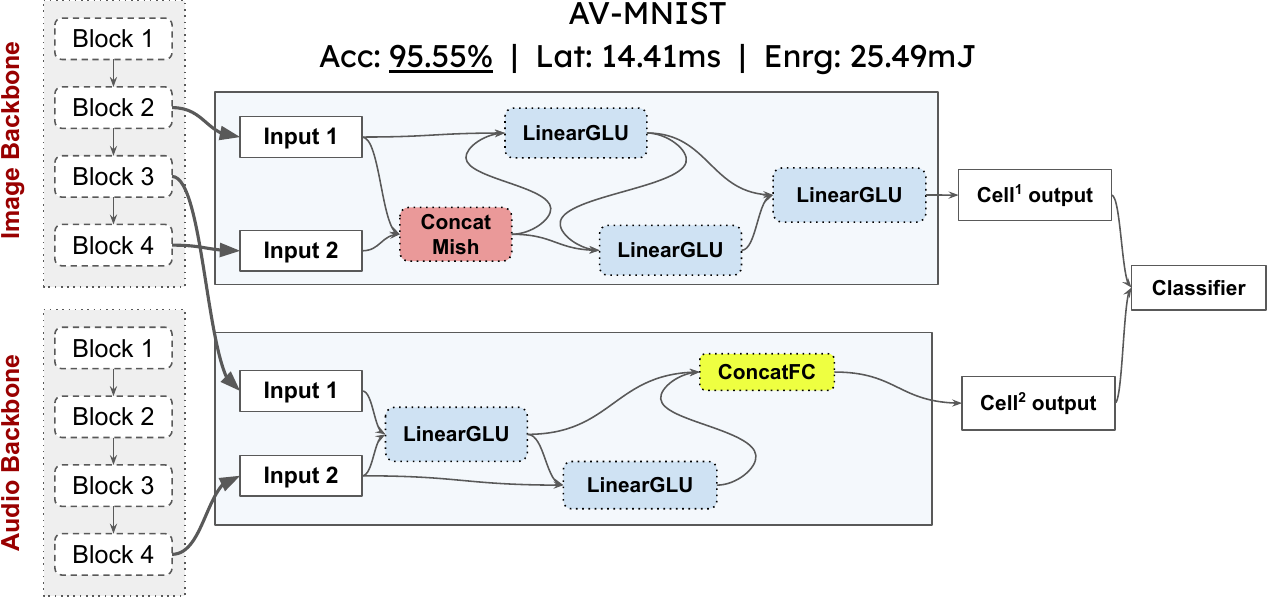}
    \caption{Visualization of the most-accurate MM-NN for the AV-MNIST dataset on the TX2.}
    \label{fig:avmnist_2}
\end{figure}

\begin{figure}[tp]
\centering
    \includegraphics[width=0.8\textwidth]{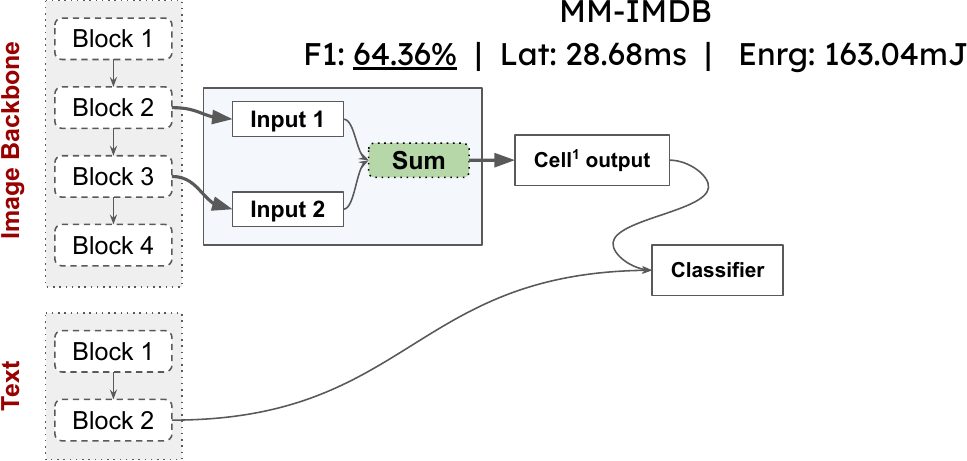}
    \caption{Visualization of the most-accurate MM-NN for the MM-IMDB dataset on the TX2.}
    \label{fig:imdb}
\end{figure}

\begin{figure}[tp]
\centering
    \includegraphics[width=0.8\textwidth]{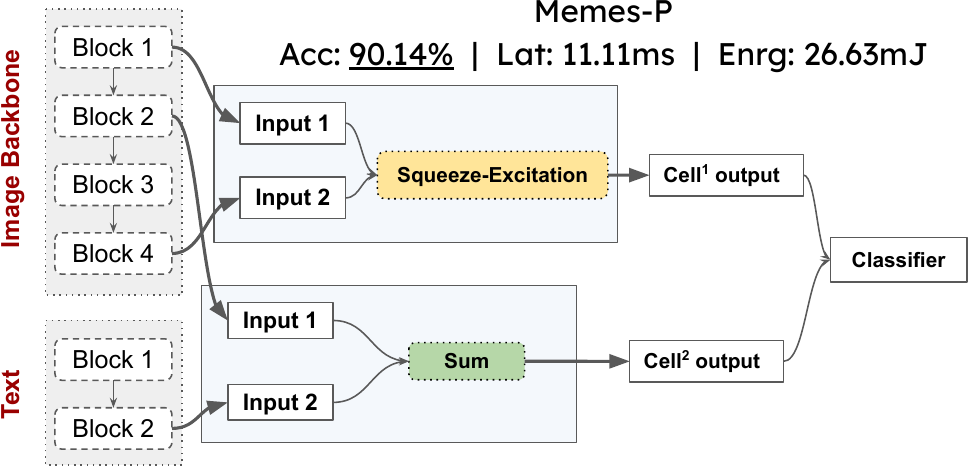}
    \caption{Visualization of the second most-accurate MM-NN for the Memes-P dataset on the TX2.}
    \label{fig:memes}
\end{figure}

\noindent As depicted in Figure \ref{fig:avmnist_1}, to achieve a latency-efficient MM-NN on the AV-MNIST dataset when deployed on the NVIDIA Jetson TX2, \textit{Harmonic-NAS} could find a tailored fusion design with less hardware demanding fusion operators. Furthermore, as reported in Table \ref{tab:pf_avmnist}, the first-stage search of \textit{Harmonic-NAS} has adapted the design of the backbones to less computationally complex and energy-demanding ones. For instance, the obtained backbones for the image and audio modalities yield high TOP-1 accuracy, resulting in rich feature joint embedding and consequently achieving higher accuracy in the context of multimodal fusion.\\

\noindent To further enhance the accuracy of the MM-NN, \textit{Harmonic-NAS} explored more intricate fusion macro architectures as shown in Figure \ref{fig:avmnist_2}. This strategic adaptation of our second-stage search engine has yielded the discovery of accurate MM-NN than SoTA baselines at the cost of increased latency and energy. These findings underscore the capacity of \textit{Harmonic-NAS} in tailoring the design of MM-NNs for resource-constrained hardware devices to adapt to different deployment scenarios and application requirements.